# Enhanced 3D brain tumor segmentation using assorted precision training

**Adwaitt Pandya[1*], Ozioma Collins Oguine[2*], Harita Bhargava[1*] and Shrikant Zade[1*]**

[1]Department of Computer Science, Oriental Institute of Science and Technology, Bhopal, India
[2]Department of Computer Science, University of Abuja, Abuja, Nigeria

[*]**Correspondence:**
Adwaitt Pandya,
adwaitt1999@gmail.com
Ozioma Collins Oguine
oziomaoguine007@gmail.com
Harita Bhargava
haritabhargava28@gmail.com
Shrikant Zade
cdzshrikant@gmail.com

A brain tumor is a medical disorder faced by individuals of all demographics. Medically, it is described as the spread of non-essential cells close to or throughout the brain. Symptoms of this ailment include headaches, seizures, and sensory changes. This research explores two main categories of brain tumors: benign and malignant. Benign spreads steadily, and malignant express growth makes it dangerous. Early identification of brain tumors is a crucial factor for the survival of patients. This research provides a state-of-the-art approach to the early identification of tumors within the brain. We implemented the SegResNet architecture, a widely adopted architecture for three-dimensional segmentation, and trained it using the automatic multi-precision method. We incorporated the dice loss function and dice metric for evaluating the model. We got a dice score of 0.84. For the tumor core, we got a dice score of 0.84; for the whole tumor, 0.90; and for the enhanced tumor, we got a score of 0.79.

## 1. Introduction

A tumor is an abnormal growth of cells in the brain that may or may not be cancerous. Tumors are generally classified into two classes: benign and malignant. Benign tumors are considered non-cancerous, i.e., they grow locally and do not spread to other tissues. They can be fatal if they develop near vital organs like the brain, even after being non-cancerous. Malignant tumors are considered cancerous. New cells are constantly produced in our body to replace the old ones; sometimes DNA gets damaged during this renewal process, so the new cells develop abnormally. These cells continue to multiply faster, thus forming a tumor. Malignant tumors can spread and affect other tissues. Tumors that affect the central nervous system are known as gliomas. Following are the constituents of gliomas:

- **Edema:** Finger-like projection, an agglomerate of fluid or water. FLAIR and T2-weighted sequences produce the best results.
- **Necrosis:** Collection of dead cells. Best seen in the T1 post-contrast sequence.
- **Enhancing tumor:** Indicates breakdown of the blood–brain barrier. Seen in T1c post-contrast sequence.
- **Non-enhancing tumor**: Seen in regions not included in edema, necrosis, or enhancing tumor.

There are many reasons why a person can have a brain tumor-like growth of cells uncontrollably in the brain due to mutation or defect in a gene. This is a reason for causing a brain tumor. Exposure to large amounts of X-rays is also an environmental cause which leads to the development of brain tumors. A tumor's effects on your body are evaluated by its size, location, and growth rate. General symptoms include changes in the pattern of headaches, nausea or vomiting, vision problems like blurred vision, tiredness, speech and hearing difficulties, and memory problems. Early diagnosis of a tumor provides the patient with the best chance of successful treatment. There are fewer chances of survival, a high cost of treatment, and many more problems arise if the care is delayed.

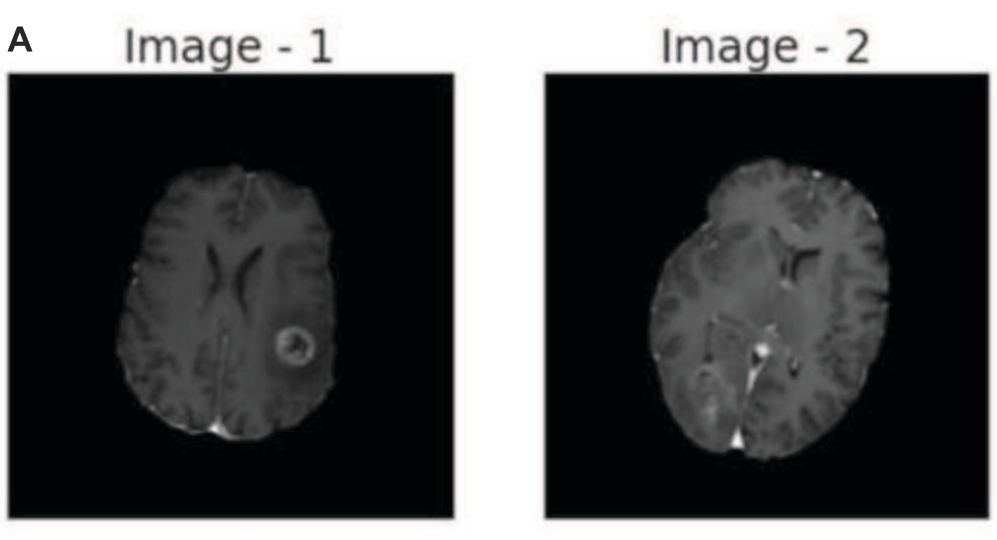

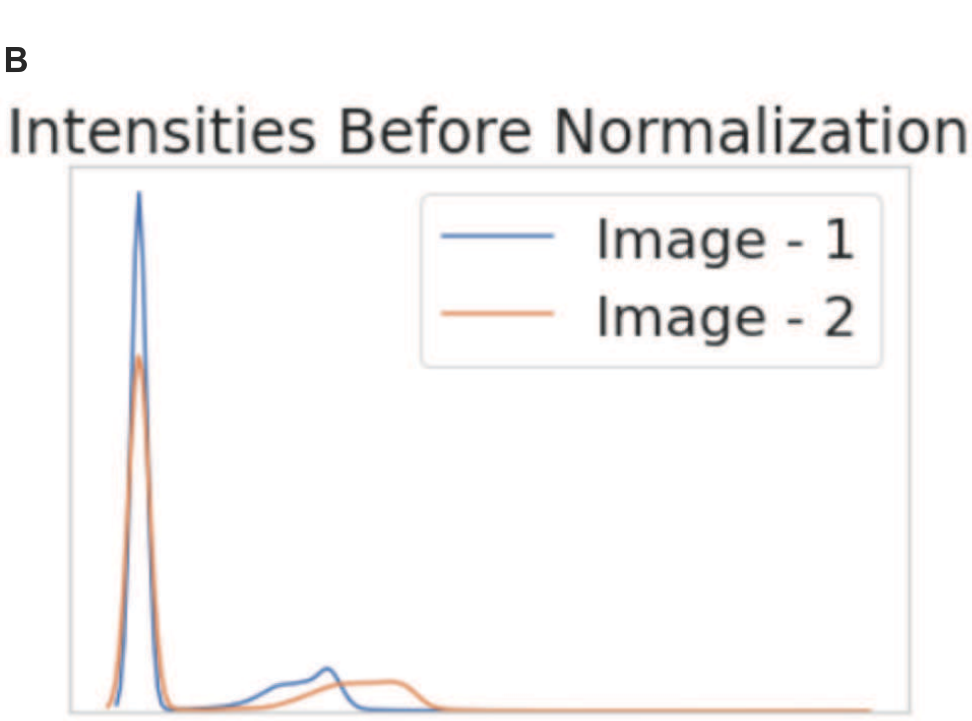

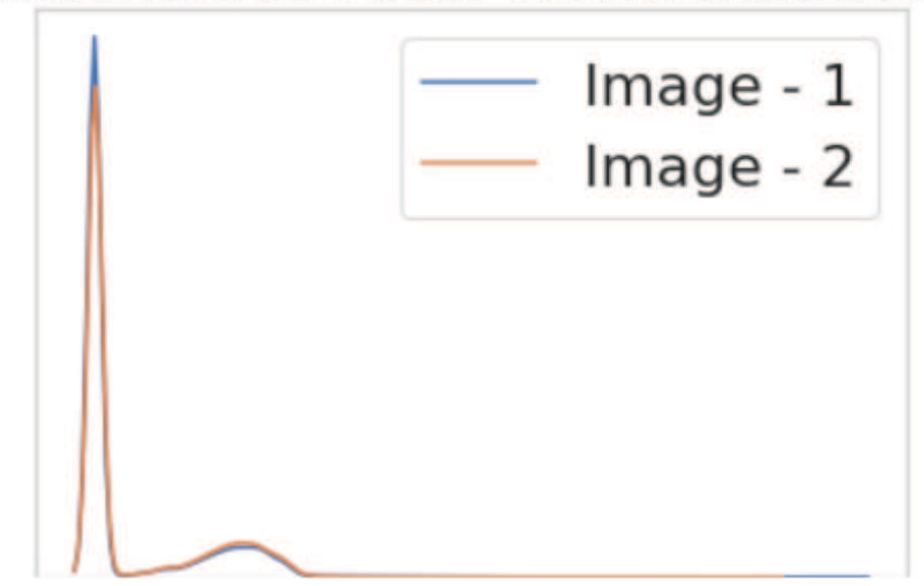

**FIGURE 1 | (A)** Two randomly selected images from the T1wCE modality. **(B)** Intensities of the two images before normalization. **(C)** Intensities of the two images after normalization. It is imperative to note that certain types of tumors are best seen in different modalities, like edema, which is best seen in T2-weighted sequences and FLAIR images. Necrosis is best visible in the T1 post-contrast sequence, and an enhancing tumor is best seen in the T1c post-contrast sequence (13).

Early diagnosis improves the outcomes of treatment. Diagnosing a brain tumor generally begins with a magnetic resonance imaging (MRI) scan. MRI is an imaging technique that utilizes magnetic fields and radio waves to create detailed images of the organs and tissues. MRI can be used to measure the tumor's size. For diagnosing a tumor, the accuracy of conventional MRI is generally satisfactory, but we should not rely heavily on it. One of the important and efficient techniques for diagnosing tumors is brain tumor segmentation. The technique of separating tumors from other brain parts in an MRI scan of the brain is called brain tumor segmentation. It separates tumors from normal brain tissues. Tumor segmentation helps in correctly identifying the spatial location of a tumor. Brain tumor segmentation proves to be useful for diagnosis and treatment planning. However, sometimes it is hard to segment the tumor because of irregular boundaries in MRI scans. If a tumor is detected on time due to segmentation, it will prove to be very convenient from the doctor's perspective to commence treatment planning as soon as possible.

Our dataset consists of a neuroimaging informatics technology initiative (NIFTI) file format. About 10 years ago, the NIFTI file format was envisioned as a replacement for the 7.5 file format for analysis. In image informatics for neuroscience and even neuroradiology research, NIFTI files are frequently employed. For our project, we used the brain tumor segmentation (BraTS) dataset. The Radiological Association of North America (RSNA), the American Society of Neuroradiology (ASNR), and the Medical Image Computing and Computer-Assisted Interventions (MICCAI) society are working together to arrange the BraTS challenge. The model used by us for segmentation is "SegResNet," and we have trained it on the BraTS 2021 (1–5) (Task 1) dataset. The following work contains a detailed description of the dataset, proposed methodology, comparative analysis, and results.

## 2. Theoretical background

### 2.1. Technology stack

We used the Kaggle notebooks for training, validation, and testing our model. By default, Kaggle provides P100 GPUs for all notebooks with 16 GB of RAM and 16 GB of storage space. We used PyTorch version 1.10.0 and MONAI 0.7.0 for coding. We also used intensity normalization (6) version 2.1.1 for normalizing intensities.

### 2.2. Dataset

We used the BraTS 2021 task-1 dataset. The collection includes segmentation masks, Native (T1), T1-weighted (T1Gd), T2-weighted (T2), and T2-Fluid Attenuated

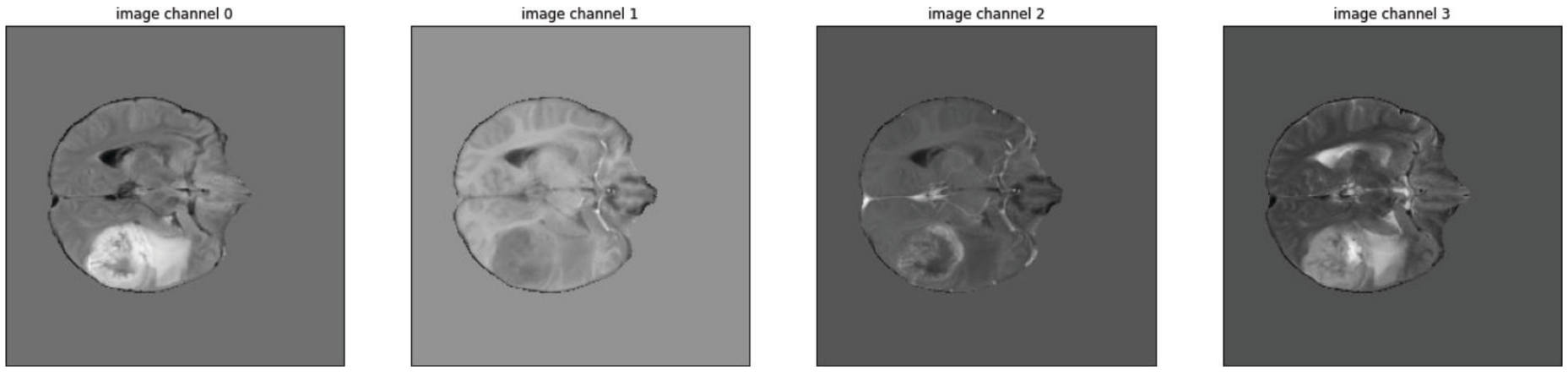

**FIGURE 2 |** Augmented samples of different modalities. Each "image channel" corresponds to a different modality.

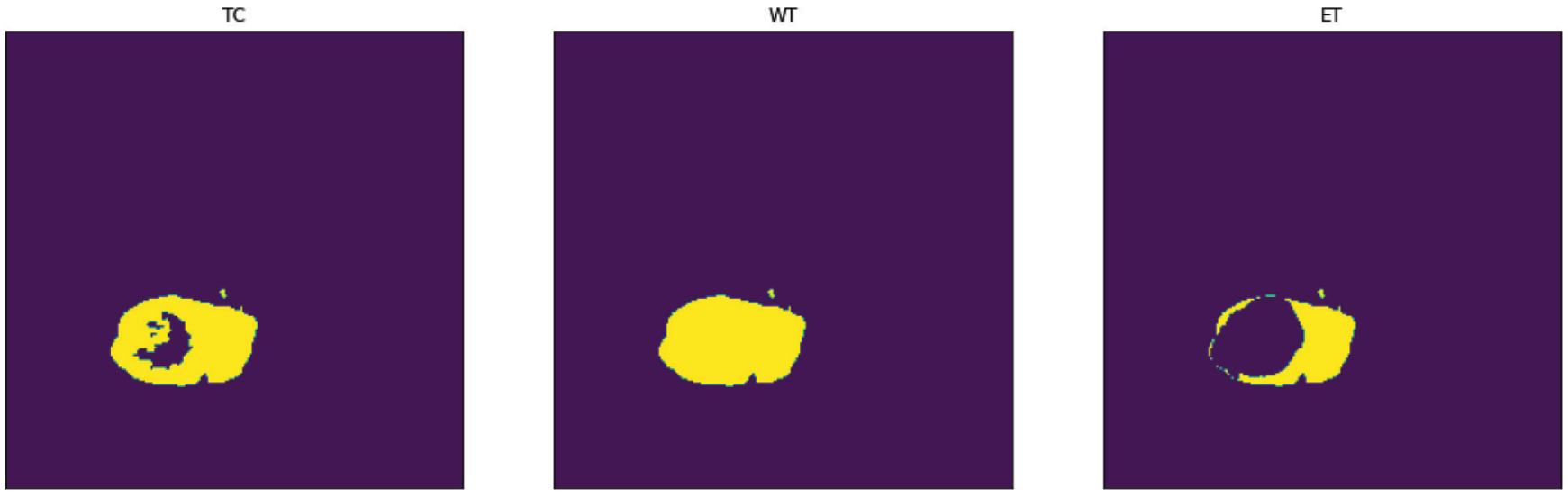

**FIGURE 3 |** Brain tumor segmentation masks.

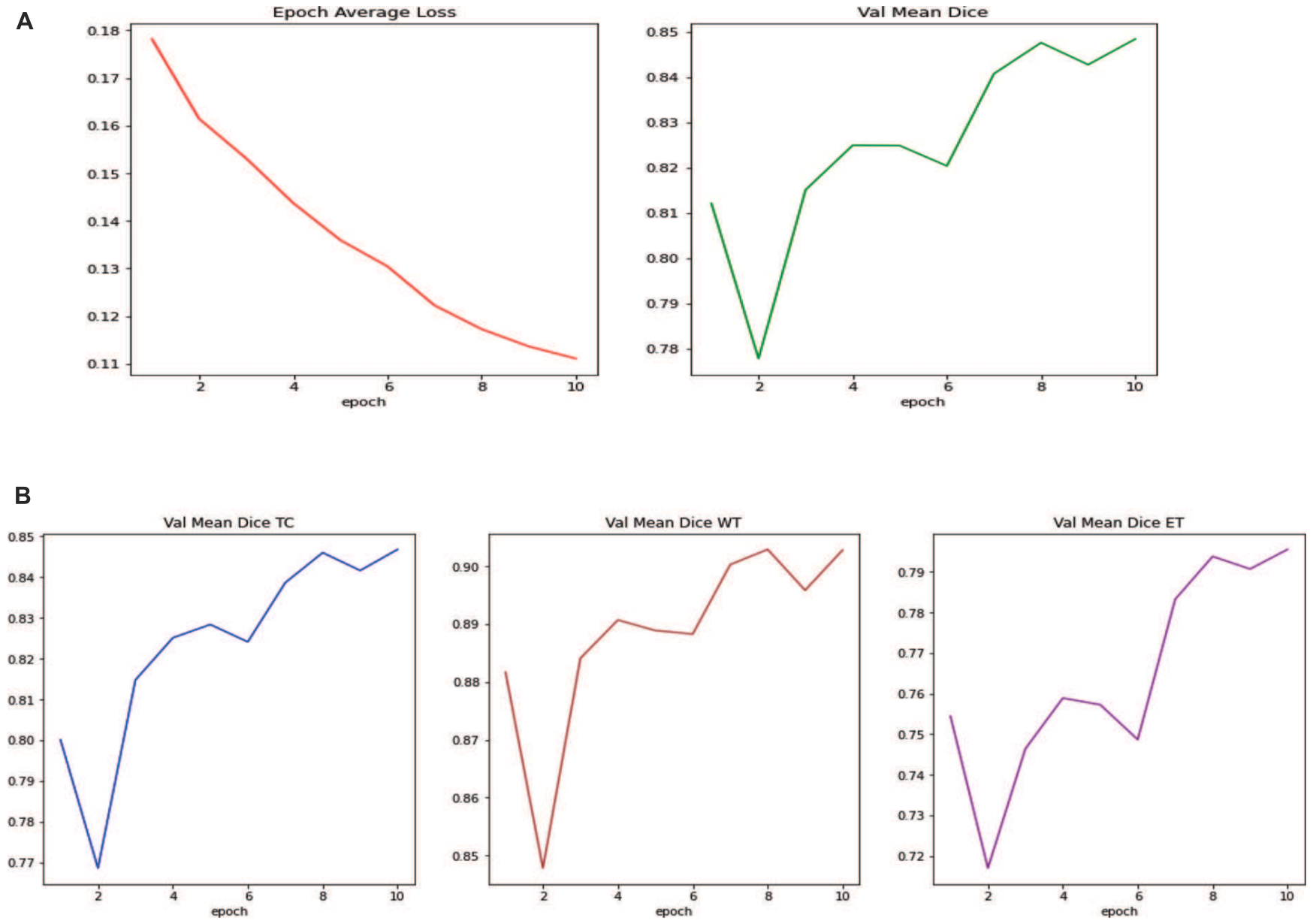

**FIGURE 4 | (A)** Average loss (left) and average dice score (right) per epoch. **(B)** Dice score for tumor core (left), whole tumor (center), and enhanced tumor (right).

Inversion Recovery (Recovery) NIFTI volumes for 1,251 individuals from various sources and in axial, sagittal, and coronal orientation. All of the files were NIFTI volumes containing 240 image slices. Each image was $240 \times 155$ in size.

## 2.3. Literature survey

For the purpose of segmenting brain tumors, Havaei et al. proposed a CNN architecture that not only utilized local and global features concurrently. They obtained dice scores of 0.85 for segmenting the entire tumor, 0.78 for segmenting the tumor core, and 0.73 for improving tumor segmentation (7).

Pereira et al. (8) explored a way to counter large spatial and structural variability by incorporating small $3 \times 3$ kernels in their proposed architecture. They attained a dice score of 0.88 on the whole tumor segmentation; for tumor core segmentation, they were able to get a score of 0.83, and for enhancing tumor, they got 0.77. Myronenko et al. (9) described an encoder-decoder-like architecture and a variational autoencoder branch. Their model yielded a dice score of 0.81, 0.90, and 0.86 on enhancing tumor, whole tumor, and tumor core segmentation.

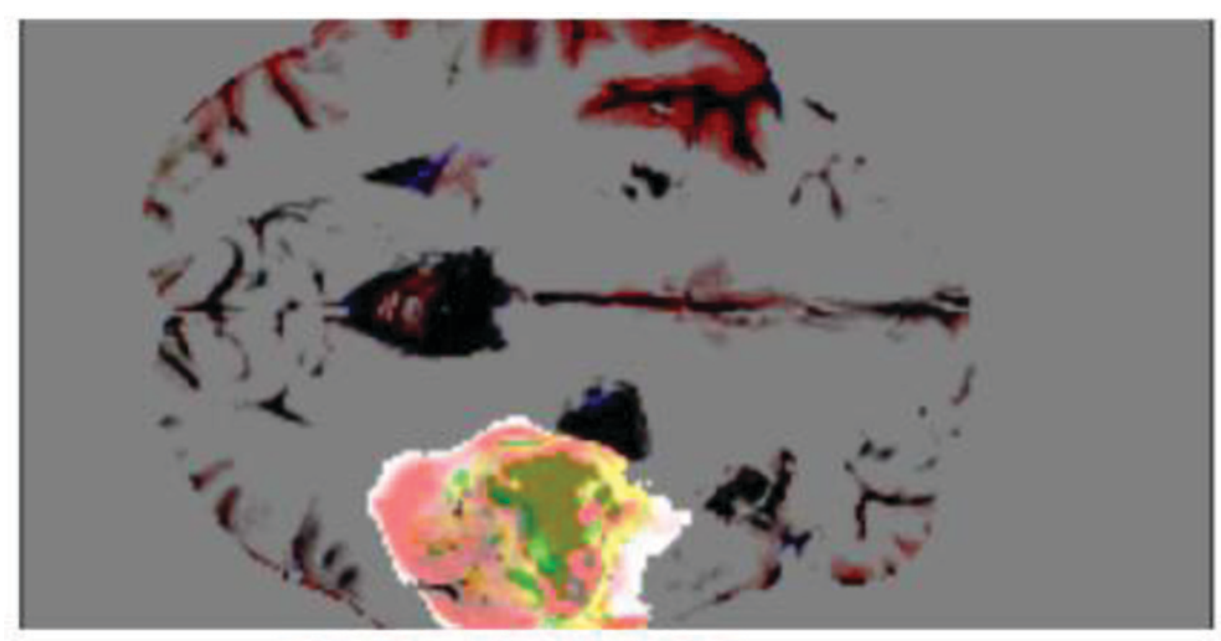

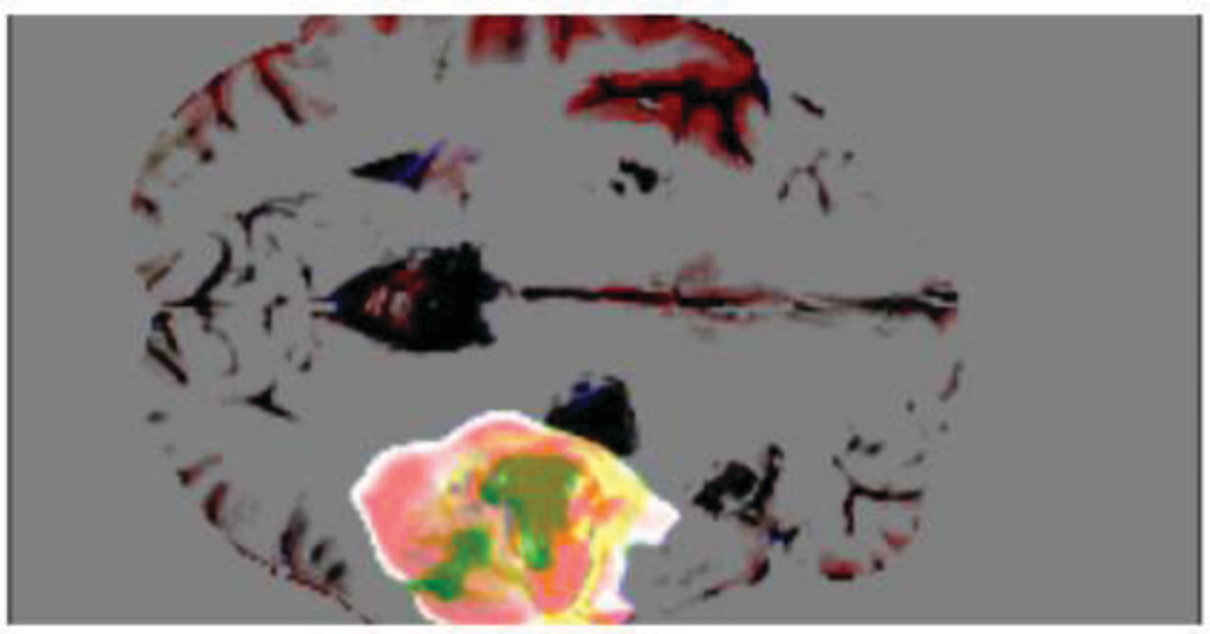

**FIGURE 5 |** The model's output **(top)** is compared to the actual tumor **(bottom)**. The yellow region is the enhanced tumor, the yellow and green regions constitute the tumor core, and the yellow, red, and green regions constitute the whole tumor.

An anisotropic and dilated convolution filter-based cascade of a fully convolutional neural network was proposed by Wang et al. (10). They broke the issue down into a series of binary categorization issues. On the entire tumor, tumor core, and improving tumor segmentation, their model received scores of 0.78, 0.87, and 0.77, respectively.

Mohammadreza et al. (11) extracted text-on-descriptor features such as histograms and first-order intensities, which were then fed into a Random Forest Classifier. They scored 0.84 on the whole tumor, 0.82 on the enhancing tumor, and 0.78 on the tumor core segmentation.

Lyu et al. (12) used a 2-stage model; for stage 1, they used an encoder-decoder-like architecture and variation autoencoder regularization. In stage 2, the network uses attention gates and is trained on a dataset formed by stage 1 output. Their dice scores for the whole tumor, tumor core, and enhancing tumor were 0.87, 0.83, and 0.82, respectively.

## 3. Methodology

**Preprocessing:** The dataset comes from various sources. We used the intensity normalization technique described by Shinohara et al. (6) over every image of every modality (FLAIR, T1w, T1wCE, and T2w) to solve the difference in intensities. Separate normalizers were used for each modality and were trained on the images belonging to their respective modalities. Since the images were in different orientations, we oriented all of them in the RAS orientation.

**Image Augmentations:** We cropped the image, keeping the region of interest of 224 × 224 × 144. We then randomly flipped the images across all three axes and randomly scaled and shifted the intensities. We used no augmentations for validation or testing.

**Loss Functions and Metric:** We used the dice loss function (14) and metric, a region-based loss function, to calculate similarities. Mathematically, the dice coefficient can be expressed as follows:

$$D = \frac{2\sum_i^N p_i g_i}{\sum_i^N p_i^2 + \sum_i^N g_i^2}$$

where $D$ is the dice coefficient, $p$ and $g$ are pairs of corresponding pixel values.(15) For training, we added a small constant to the denominator to tackle the scenarios in which the denominator becomes less than 0.

**Hyperparameters:** We initialized the model with 16 filters, keeping the number of input channels equal to 4 and yielding an output of 3 channels, corresponding to the three classes discussed earlier. We also applied a dropout with a dropout rate of 0.2. Due to memory constraints, we kept the batch size to 1. We used the Adam optimizer for training with a learning rate of 0.0001. We applied L2 regularization with a regularization coefficient set to 0.00001 and trained the model for 10 epochs.

**Model Architecture:** We used the SegResNet architecture with the number of downsampling blocks in each layer being 1, 2, 2, and 4, respectively. The number of upsampling blocks in each layer is 1, 1, and 1, respectively.

**Training:** We trained our model for 10 epochs and saved the best-performing model. We modified the traditional training technique and used the mixed-precision (16) training method, which enhances performance and efficiency and reduces memory requirements. We used automatic mixed precision for both the training and validation tasks.

**TABLE 1 |** Comparison of dice scores obtained by different methods.

| Researches | Number of test images | Dice scores |
|---|---|---|
| Havaei et al. (7) | 200 2d slices and approximately 6,000 2D images | WT = 0.85, TC = 0.78, ET = 0.73 |
| Pereira et al. (8) | The training set contains 20 HGG and 10 LGG. | WT = 0.88, TC = 0.83, ET = 0.77 |
| Myronenko et al. (9) | Training dataset included 285 cases (210 HGG and 75 LGG). | ET = 0.81, WT = 0.90, CT = 0.86 |
| Wang et al. (10) | The training set contains images from 285 patients (210 HGG and 75 LGG). | WT = 0.7831, CT = 0.8739, ET = 0.7748 |
| Soltaninejad et al. (11) | The dataset was tested on 11 multimodal images and the BRATS 2013 clinical dataset using 30 multimodal images. | WT = 0.80, TC = 0.89 |
| Lyu et al. (12) | The BraTS 2020 dataset containing 259 HGG and 110 LGG cases | ET = 0.79, WT = 0.90, TC = 0.83 |
| Proposed | 1251 NIFTI volumes of 240 image slices each | ET = 0.71, WT = 0.90, TC = 0.84 |

## 4. Result analysis

We converged our loss to 0.11 and got an average dice score of 0.84 on the validation set. As for the separate classes, our dice scores were as follows (on the validation set):

TC = 0.84
WT = 0.90
ET = 0.79

On the test set, we got a mean dice score of 0.86; in the TC class, the dice score was 0.86; in the WT class, it was 0.92; and finally, on ET, it was 0.81.

## 5. Discussion

Brain tumor segmentation proves to be an effective tool for accurately diagnosing the tumor and its constituents. Our model can segment a tumor from an MRI image efficiently. The model was trained on 750 NIFTI volumes and validated and tested on 250 NIFTI volumes. We were able to get our loss down to 0.11 and got a mean dice score of 0.84. We believe that more data could significantly improve the model's performance for future research.

## 6. Conclusion

In this study, we used the SegResNet architecture to segment the brain tumor. Our model produces great results on 200 test cases. Our model's best score produced a dice score of 0.86 on TC, 0.92 on WT, and 0.81 on ET. In contrast, the mean dice score was 0.84.

## Disclosure

The authors declare that they have no competing interests with anyone in publishing this manuscript.

## Author contributions

All authors made substantial contributions in conscripting the manuscript and revising it critically for important intellectual content, agreeing to submit it to the current journal, and final approval of the version.